\pdfoutput=1

\documentclass[11pt]{article}

\usepackage{authblk}

\usepackage{acl}

\usepackage{times}
\usepackage{latexsym}

\usepackage[T1]{fontenc}

\usepackage[utf8]{inputenc}

\usepackage{microtype}

\usepackage{inconsolata}

\usepackage{lipsum}
\usepackage{yhmath}
\usepackage{amsfonts}\usepackage{hyperref}

\usepackage{amsmath}

\usepackage{graphicx}
\graphicspath{ {.} }
\usepackage{calligra}
\usepackage{physics}
\usepackage{listings}
\usepackage{xcolor}
\usepackage{colortbl}
\usepackage{blindtext}
\usepackage{tikz}
\usepackage{caption}
\usepackage{subcaption} 
\usepackage{arydshln}
\usepackage{booktabs}
\definecolor{LightGray}{rgb}{0.93,0.93,0.93}
\definecolor{Fuchsia}{rgb}{0.6,0.0,0.5}
\definecolor{StrongGreen}{rgb}{0.0,0.7,0.1}
\definecolor{StrongRed}{rgb}{0.7,0.1,0.0}
\usepackage[colorinlistoftodos,disable]{todonotes}
\usepackage{enumitem}
\usepackage{float}
\usepackage[normalem]{ulem}
\usepackage{calc}

\newcommand\Ccancel{\bgroup\markoverwith
{\textcolor{red}{\rule[+0.5ex]{2pt}{0.8pt}}}\ULon}
\newcommand\Cu{\bgroup\markoverwith
{\textcolor{orange}{\rule[-0.5ex]{2pt}{0.8pt}}}\ULon}

\newcommand{\tn}[1]{\scriptsize#1\normalsize}

\newcommand{\maxsmalltodo}[1]{\todo[size=\tiny]{\textcolor{white}{TODO: }#1}}

\newcommand{\fytodom}[1]{\todo[color=red,author=FY,list]{#1}}

\newcommand{\fytodo}[1]{\fytodom{#1}}
\newcommand{\fydone}[1]{\todo[color=green,author=FY,nolist]{#1}} 
\renewcommand{\fydone}[1]{} 
\newcommand{\src}{\ensuremath{\mathbf{x}}}

\newcommand{\density}{\text{density}}

\DeclareMathOperator{\simi}{sim}

\DeclareMathOperator{\lcs}{lcs}
\DeclareMathOperator{\ecs}{ecs}

\newcommand{\mlevt}[1]{\texttt{TM$^{#1}$-LevT}}
\newcommand{\levt}[0]{\texttt{LevT}}

\newcommand{\tmlevt}[0]{\texttt{TM-LevT}}
\newcommand{\Lev}[0]{\text{LED}}
\title{A Systematic Comparison of Retrieval Techniques for Retrieval Augmented Neural Machine Translation}
\title{A Systematic Comparison of Retrieval Techniques for Neural Machine Translation with Translation Memories}
\title{Retrieving Examples from Translation Memories in NMT: \\A Systematic Study}
\title{Retrieving Examples from Memory for Retrieval Augmented Neural Machine Translation: A Systematic Comparison}

\newcommand{\systran}{\ensuremath{\clubsuit}} 
\newcommand{\isir}{\ensuremath{\heartsuit}} 
\newcommand{\CAT}{Computer Aided Translation\xspace}

\author[$\isir$,$\systran$]{Maxime Bouthors}
\author[$\systran$]{Josep Crego}
\author[$\isir$]{Fran{\c c}ois Yvon}
\affil[$\isir$]{Sorbonne Université, CNRS, ISIR, F-75~005 Paris, France}
\affil[$ $]{\texttt{lastname@isir.upmc..fr}}
\affil[$\systran$]{ChapsVision, 4 rue du Port aux Vins, F-92~150 Suresnes, France}
\affil[$ $]{\texttt{\{mbouthors,jcrego\}@chapsvision.com}}

\begin{document}
\maketitle

\begin{abstract}
	Retrieval-Augmented Neural Machine Translation (RAMT) architectures retrieve examples from memory to guide the generation process. While most works in this trend explore new ways to exploit the retrieved examples, the upstream retrieval step is mostly unexplored. In this paper, we study the effect of varying retrieval methods for several translation architectures, to better understand the interplay between these two processes.
	We conduct experiments in two language pairs in a multi-domain setting and consider several downstream architectures based on a standard autoregressive model, an edit-based model, and a large language model with in-context learning. Our experiments show that the choice of the retrieval technique impacts the translation scores, with variance across architectures. We also discuss the effects of increasing the number and diversity of examples, which are mostly positive across the board.
\end{abstract}

\section{Introduction \label{sec:introduction}}
\fydone{Fix references in the intro}
Retrieval-Augmented Language Models and Translation Models are getting a lot of traction (see \cite{li-etal-2023-survey} for a recent review).
For translation tasks, the use of retrieval-based techniques that identify the most relevant segment(s) in a Translation Memory (TM) has long been used in professional \CAT{} environments \cite[]{bowker-2002-computer}, where the retrieved segments provide translators with valuable suggestions. Segments closely resembling the source sentence to be translated can also be directly edited to speed up translation. Such ideas have also been used in Machine Translation (MT), first in the example-based tradition
\cite{nagao-1984-framework,somers-1999-review,carl-etal-2004-recent}, then in the statistical-based paradigm \cite[]{koehn-senellart-2010-convergence}, more recently for neural machine translation (NMT).

There are several ways to take advantage of translation examples in NMT architectures: \citet{farajian-etal-2017-multi} use a small set of examples to perform on-the-fly, lightweight, fine-tuning (using both source and target sides); \citet{bulte-tezcan-2019-neural} simply concatenate the (target side of) a handful of examples on the source side of the encoder, leaving the rest of their autoregressive decoder unchanged; \citet{xu-etal-2023-integrating} repurpose the edit-based architecture of \citet{gu-etal-2019-levenshtein} to compute new translations from existing ones with a non-auto-regressive (NAT) decoder; finally, in-context learning (ICL) in large language models (LLMs) provides yet another way to seamlessly combine TMs with text generation (see \citet{moslem-etal-2023-adaptive}, \textsl{inter alia}).

These studies (and several others, fully discussed in \textsection\ref{sec:related}) not only differ in the way they use examples but also in the way TMs are searched, retrieved, and filtered. This makes the direct comparison between these proposals sometimes difficult to reproduce and analyze. Furthermore, it also prevents precisely assessing the computational complexity of the complete translation pipeline.

In this paper, we perform experiments with several representative retrieval methods that we systematically combine with multiple RAMT architectures. In doing so, our main goal is not to compare these downstream architectures, but rather to better understand the interplay between the retrieval and generation tasks, to make the trade-offs between these steps explicit, and to formulate recommendations regarding future uses of TMs in NMT. 
\fydone{Rephrase, may be}

Specifically, we address the following questions:\fydone{Add some conclusions at the end}
\begin{itemize}
	\item How much does the example selection impact translation performance? Is one retrieval technique always better than the others, irrespective of the MT architecture, or is it necessary to adapt the former to the latter?
	\item Do we need multiple examples? If so, what makes a good set of examples? Does the diversity of examples help?
	\item By restricting retrieval to a restricted subset of examples based on the source domain, can we expect better performances, even though the quality of the best retrieved match is decreased?
\end{itemize}
We notably find that (a) retrieval actually matters for edit-based and in-context learning; (b) existing retrieval pipelines can be simplified at inference; (c) optimizing source coverage and/or instance diversity is helping, especially when the closest match is poor.

\section{Retrieval in NMT architectures \label{sec:architectures}}

\subsection{Retrieval Pipeline}
\label{ssec:retrieval-pipeline}

\begin{figure*}[htpb]
	\centering
	\includegraphics[width=\textwidth]{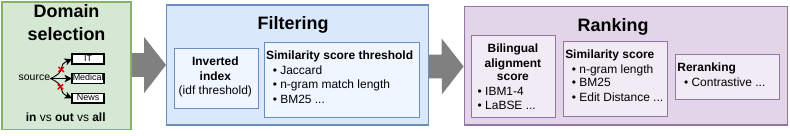}
	\caption{\label{fig:retrieval-pipeline} High-level overview of the retrieval pipeline in fuzzy-matching.}
\end{figure*}

In the information retrieval (IR) framework, given a query $q$, a set of documents $D$ is filtered and/or ranked according to a retrieval score. The challenge is to craft a score $s(q, d)$ that will retrieve documents $d$ that are relevant for a downstream task. In MT, $q$ is a source sentence, and $D$ is a translation memory from which we can extract relevant (source, target) pairs ($d=(x,y)$).
The retrieval process can be divided into three steps (Figure~\ref{fig:retrieval-pipeline}):
\begin{enumerate}

	\item \textbf{Domain selection} selects the corpus to retrieve examples from, typically based on domain/genre similarity;
	\item \textbf{Filtering} narrows down the set of relevant examples based on superficial comparison between the source query $q$ and each example $d$.
	      \begin{itemize}
		      \item Filtering can use a simple similarity score to filter TM candidates based on some minimal threshold: \citet{bulte-tezcan-2019-neural} uses Jaccard similarity between bag-of-word representations; \citep{xu-etal-2020-boosting,bouthors-etal-2023-towards} use an n-gram match similarity;

		      \item Filtering can also be controlled by the specification of the indexing vocabulary, which typically excludes frequent words, thereby shortening the list of similar documents (see Appendix~\ref{appendix:bm25}).
	      \end{itemize}
	\item \textbf{Ranking} uses a retrieval score such as n-gram overlap \citep{xu-etal-2020-boosting}, BM25 score \citep{gu-etal-2018-search,cheng-etal-2022-neural}, edit distance (ED) \citep{bulte-tezcan-2019-neural,xu-etal-2020-boosting,bouthors-etal-2023-towards}, cosine similarity between $q$ and $x$'s embeddings \cite{xu-etal-2020-boosting,pham-etal-2020-priming,vilar-etal-2023-prompting}. \emph{Incremental ranking} or \emph{Weighted Coverage} can also be used to enforce diversity when retrieving multiple samples \citep{cheng-etal-2022-neural,agrawal-etal-2023-context,sia-duh-2023-context}.
\end{enumerate} \fydone{Clarify this - retrieval vs. ranking}
In a final \textbf{selection step}, only the top-$k$ most similar candidates are eventually retained. Depending on the choice of the filtering parameters, the actual number of examples retrieved for a given query may be strictly smaller than $k$. For some domains, retrieval may even return an empty list. 

\subsection{Measuring Retrieval Quality}
\label{ssec:retrieval-quality}
The effects of a retrieval strategy are only observed once a translation model is trained and evaluated. As this is an excessively costly process, we introduce several aspects that define \textsl{a priori} the quality of retrieved similar examples. For a single example $d$, this includes its semantic relatedness with $q$ or the lexical overlap with the query; $d$'s length also matters, as long examples may include irrelevant words that can hurt translation \citep{xu-etal-2020-boosting}, and increase the computational processing cost. Now, looking at \emph{sets of examples}, we would also like them to cover most query words, while remaining diverse and short on average.

To evaluate these facets, we compute the following scores for each set of similar examples:
\begin{itemize}
	\item \textbf{Coverage} is the proportion of query tokens covered by the example tokens. It can be defined in several ways (bag-of-word recall, modified recall\footnote{Each source token can only be covered at most once.}, n-way alignment score\footnote{Based on an alignment graph between examples and query that forbids swapping, as defined by \citet{bouthors-etal-2023-towards}.}).
	\item \textbf{Relevance} is the proportion of contributing tokens from the example tokens (with the same three underlying definitions as coverage). All other words are deemed lexically irrelevant; we report an average over examples.
	\item \textbf{Length} is the average number of tokens of retrieved examples. 
	      \fydone{Diversity. Explain - BLEU scores ??}
\end{itemize}
In the next paragraphs, we describe how these quantities are controlled during retrieval.

\subsection{Smoothed Longest Common Subsequence}
\label{ssec:smoothed-lcs}

The Levenshtein edit distance (LED) is widely used as the ranking function in RAMT (see \textsection\ref{ssec:retrieval-pipeline}). It counts the minimum number $\Delta(x, q)$ of word-level operations (deletion, insertion, replacement) required to edit $x$ into $q$ and then normalizes:
\begin{equation}
	\Lev(x, q) = 1 - \frac{\Delta (x, q)}{\max (|x|, |q|)},
	\label{eq:lev-dist}
\end{equation}
Using $\Lev$ mainly selects examples that are lexically similar to the source. As it penalizes non-matching parts (in the normalizer), it may discard long examples containing good matches in a substring. Such examples can still be relevant if they yield a high coverage of the query.

As an alternative to LED, setting the deletion cost to zero in~\eqref{eq:lev-dist} computes the Longest Common Subsequence (LCS) between $x$ and $q$, which maximizes the coverage at the expense of relevance. This also means that there is no penalization for length, which can lead to long and hard-to-exploit examples. We propose a smoothed version, namely $\delta$-LCS, with a small non-zero deletion cost $\delta$. $\delta$-LCS thus performs a trade-off between coverage, relevance, and length. Details are in Appendix~\ref{appendix:d-lcs}.


\subsection{Controlling Diversity: Contrastive Retrieval}
\label{ssec:retrieval-diversity}

As is well known in the IR literature, it is unproductive to retrieve multiple identical examples. Diversity can help increase coverage without hurting the relevance of individual examples. For RAMT, a small number of diversity preserving approaches have been proposed: \citet{cheng-etal-2022-neural} (Levenshtein distance), \citet{agrawal-etal-2023-context} (n-gram overlap) and \citet{sia-and-duh-2023-incontext} (BM25) rely on an iterative algorithm inspired by the Maximal Marginal Relevance (MMR) criterion of \citet{goldstein-carbonell-1998-summarization}. In a nutshell, this means that the ranking scoring function is iteratively updated to downgrade candidate examples that are either too similar to already selected examples or that cover already covered words. In our experiments, we follow \citet{cheng-etal-2022-neural}, and after selecting $|M|$ matches in $M$, we penalize the ranking scores of remaining candidates with the following term:
\begin{equation}
	\frac{\alpha}{|M|} \sum_{x \in M} \Lev (\cdot, x),\label{eq:contrast}
\end{equation}
where $\alpha >0$ controls the strength of the penalty.


\subsection{Integrating TMs in Translation}
\label{ssec:integrating-retrieval}

In our comparisons, we consider three NMT architectures with variants. The first, called Neural Fuzzy Augmentation (NFA), implements the autoregressive approach of \citet{bulte-tezcan-2019-neural}, with minor variants. The second, \tmlevt{}, is edit-based and mostly follows \citet{xu-etal-2023-integrating}, as recently extended by \citet{bouthors-etal-2023-towards} to handle multiple matches. The third is based on in-context learning (ICL) with large LMs, using the causal BLOOM LM \cite{lescao2022bloom} and the HuggingFace Transformer library\footnote{\url{https://github.com/huggingface/transformers}.} to run the experiments. We provide full details regarding these architectures in Appendix~\ref{appendix:models-settings}.
\fydone{Could expand a bit, providing extra space - especially between edit-based and not edit based}
At a high level, the most important distinction is between the \emph{autoregressive generative approaches} of NFA and ICL, which both use an enriched context comprising the source sentence and additional source and/or target matches, and \tmlevt{}, which tries to reuse, via an editing process, subparts of the retrieved matches. Another major difference is between ICL, which inputs the source side of matches for decoding, and the other two approaches, which do not need it during generation. This notably impacts the computational cost of encoding the context.

\section{Data and Metrics \label{sec:protocols}}

\subsection{Data \label{ssec:data}}
\fydone{Mostly from mlevT paper, should rephrase a bit}

We consider two translation directions, from English to French (en-fr) and from German to English (de-en), and experiment with multiple domains.
This allows us to study a wide range of settings, with a varying density of matches: our datasets include ECB, EMEA, Europarl, GNOME, JRC-Acquis, KDE4, PHP, Ubuntu, OpenSubtitles, and Koran\footnote{These data can be downloaded from the OPUS website (\url{https://www.opus.nlpl.eu}) \citep{tiedemann-2012-parallel}.} (statistics in Tables~\ref{table:data-stats} and \ref{table:data-stats-de-en}). For en-fr, these datasets reproduce the setting of \cite{xu-etal-2020-boosting}. \footnote{Splits from \url{https://github.com/jitao-xu/tm-levt}.} For de-en, we reuse the data prepared by \citet{koehn-knowles-2017-six} with the split of \citet{aharoni-goldberg-2020-unsupervised}.\footnote{This is the \textit{test-de} test set.} The most favorable situation is to translate in a narrow domain with large TMs, ensuring that multiple good matches can be found  (e.g.\ JRC-Acquis and EMEA). However, in a narrow domain, the TM can sometimes be small (e.g., Ubuntu), and this is where other related domains can also help. On the other end of this spectrum, Europarl or News-Commentary are thematically very diverse and good matches much harder to find.

For each dataset $\mathcal D$, we compute a \textit{density} score based on the number of connected components (NCC) in a similarity graph $\Gamma$. Two translation examples are linked in $\Gamma$ if their similarity score~\eqref{eq:lev-dist} is greater than $0.4$:
\begin{equation}
	\density(\mathcal D) = 1 - \frac{1 - \text{NCC}(\Gamma)}{1 - |\mathcal D|}
	\label{eq:density}
\end{equation}
In high-density domains,  it is thus easier to retrieve relevant translation examples (see Tables~\ref{table:data-stats} and \ref{table:data-stats-de-en}).

Note that these data are not ideal. First, for some domains, the corresponding data may be included in the very large corpora used to train LLMs.
In our experiments with BLOOM, which is trained on the ROOTS corpus \cite{laurencon2022bigscience}, this is the case for JRC-Acquis, Wikipedia, Europarl, TEDTalks (en-fr).\footnote{For these domains, the ICL scores have been disregarded.}\fydone{Are we sure about JRC aquis?}

Furthermore, the en-fr test sets have been selected based on the existence of at least one close example in the same domain, using the standard LED to compute similarities. More precisely,\fydone{Check reference} the 1,000 instances in \emph{test-0.6} always have at least one match with similarity greater than $0.6$, for \emph{test-0.4} the nearest match has a similarity comprised between $0.4$ and $0.6$ (details in \cite{xu-etal-2020-boosting}).\footnote{As we use our own reimplementation of edit distances, we have observed rare cases where these conditions were not exactly met.}
\fydone{This is a retrieval issue, not a test issue}

This design allows us to focus on the effect of retrieval quality (medium vs high-scoring matches) on translation scores. It however yields absolute scores that do not compare with what would be obtained with a fully randomized selection process. For a more realistic evaluation, we use the de-en data, which, however, cover fewer domains.\fydone{de-en in the table}
In general, our experiments are more thourough with en-fr data as this language pair was used to select a subset of interesting configurations to be then also tested for de-en.


\begin{table*}[ht]
	\centering
	\scalebox{0.9}{
		\begin{tabular}{lrrrrrrrrrrrr}
			\hline
			\rowcolor{LightGray}
			domain     & ECB  & EME  & Epp  & GNO   & JRC   & KDE   & News  & PHP   & TED   & Ubu   & Wiki  & \textbf{all} \\ \hline
			size       & 195k & 373k & 2.0M & 55k   & 503k  & 180k  & 151k  & 16k   & 159k  & 9k    & 803k  & 4.4M         \\ \hdashline
			avg length & 29.2 & 16.7 & 26.6 & 9.4   & 28.8  & 10.5  & 26.4  & 14.5  & 17.7  & 5.2   & 19.6  & 18.6         \\ \hdashline
			density \% & 87.1 & 96.9 & 49.3 & 85.96 & 86.76 & 84.18 & 11.60 & 63.59 & 53.78 & 16.89 & 55.25 & 62.8
		\end{tabular}
	}
	\caption{\label{table:data-stats}Number of samples, average length of tokenized sentences, density for training sets (en-fr).}
\end{table*}

\begin{table}[ht]
	\centering
	\scalebox{0.8}{
		\begin{tabular}{lrrrrrr}
			\hline
			\rowcolor{LightGray}
			domain      & it   & kor  & law  & med  & sub  & \textbf{all} \\ \hline
			size        & 223k & 18k  & 467k & 248k & 500k & 1.46M        \\
			mean length & 9.6  & 20.4 & 28.0 & 15.5 & 8.1  & 16.3         \\
			density \%  & 53.2 & 51.4 & 58.6 & 69.5 & 74.2 & 61.4
		\end{tabular}
	}
	\caption{\label{table:data-stats-de-en}Number of samples, average length of tokenized sentences, density for training sets (de-en).}
\end{table}

As a last word of caution,
we observe that some domains are much easier to translate than others: JRC-Acquis
is very repetitive, which yields BLEU scores in the high 70's; NewsCommentary, on the other hand
barely achieves BLEU scores higher than 20. Averaged results should be looked at with care - only per-domain scores can tell the full story (Appendix~\ref{appendix:per-domain}).

\subsection{Metrics \label{ssec:metrics}}
\fydone{The trickyness of comparisons also bc the length increases with more examples, meaning smaller batches}

We report BLEU scores \citep{papineni-etal-2002-bleu} computed by SacreBLEU \citep{post-2018-call},\footnote{signature: \texttt{nrefs:1|case:mixed|eff:no|tok:13a| smooth:exp|version:2.1.0};}
as well as COMET-22\footnote{\texttt{Unbabel/wmt22-comet-da}} scores \citep{rei-etal-2022-comet} using the official implementation. Additionally, we use the multi-reference sentence BLEU scores between the target side of examples, and the translation output\footnote{Brevity penalty (BP) is removed.}, averaged over corpora, to evaluate the \emph{copy rate} of systems, i.e.\ their ability to recopy subparts of the retrieved examples.


\subsection{Implementation and Parameters}
We use in-house, open-source\footnote{\href{https://github.com/SYSTRAN/fuzzy-match}{https://github.com/SYSTRAN/fuzzy-match}} libraries that implement the various retrieval methods explored in this paper. Details regarding parameter settings are in Appendix~\ref{appendix:fm-settings}. For translation architectures, refer to Appendix~\ref{appendix:models-settings}.

In our experiments, we contrast three strategies for domain selection: in-domain, out-of-domain, no-selection. Regarding filtering (step~2 in Figure~\ref{fig:retrieval-pipeline}), we compare n-gram matching (NGM), BM25, and no filter. NGM filters out examples unless they share a common n-gram $g$ with the source $q$ of relative length greater than a threshold $\tau$ (e.g.\, $\frac{|g|}{|q|} \geq \tau$). As for BM25, we only retain the $L$ best BM25 candidates.

%

Finally, regarding ranking, we compare various definitions of the edit distance (ED) (see \textsection\ref{ssec:smoothed-lcs}) with BM25.

\paragraph{Computational issues} In our experiments below, we mostly analyze retrieval results. However, note that each retrieval pipeline yields specific training and inference computational costs. Domain selection always speeds up the subsequent steps, with a very strong impact when the target domains are small. Regarding filtering, NGM has an algorithmic complexity $O(\bar \ell \log (n \bar \ell))$ for a single query using suffix array -- with $\bar \ell$ the average sentence length and $n$ the TM size -- whereas BM25's complexity is $O(n |q|)$. Finally, regarding ranking, ED calculation takes ($O(n \bar \ell |q|)$), so again, linear w.r.t.\ $n$ for one query.

\section{Experiments \label{sec:experiments}}

\subsection{Comparing Retrieval Techniques \label{ssec:compare-IR}}
\fydone{Also speed}
We first measure how much a change in the retrieval technique actually affects the instances that are eventually retrieved. For this, we compute bag-of-word coverage, relevance, and length (introduced in \textsection\ref{ssec:retrieval-quality}). We compare a baseline NGM filter using the LED ranker, as used in \citet{bouthors-etal-2023-towards}, with filter-free pipelines.
The corresponding results for all testsets are in Table~\ref{tab:main-retrieval-avg-scores}. LED yields the highest relevance, while $\delta$-LCS and contrastive ranking yield a higher coverage. Overall, changing the retrieval technique does impact the set of instances that are used in training and inference.

\begin{table*}
	\centering\scriptsize
	\begin{tabular}{|l|*{7}{c|}} \hline
		filter    & NGM ($\tau=0.3$)          & -                                           & -                                                   & -                & -                                           & -                                          & -                                          \\
		ranker    & LED                       & LED                                         & LED                                                 & $\delta$-LCS     & $\delta$-LCS                                & BM25                                       & BM25                                       \\
		Contrast  & -                         & -                                           & $\alpha=0.3$                                        & -                & $\alpha=0.3$                                & -                                          & $\alpha=0.3$                               \\ \hline\hline
		          &                           &                                             &                                                     &                  &                                             &                                            &                                            \\[-1.9ex]
		coverage  & 48.1~~65.6~~38.1          & 58.2~~68.9~~57.0                            & 60.8~~70.5~~\makebox[\widthof{00.0}][c]{-}          & 63.8~~71.9~~62.5 & \textbf{65.8}~~\textbf{73.5}~~\textbf{64.4} & 62.2~~70.0~~\makebox[\widthof{00.0}][c]{-} & 62.3~~70.0~~\makebox[\widthof{00.0}][c]{-} \\
		relevance & 40.3~~\textbf{53.5}~~31.4 & \textbf{44.7}~~\textbf{53.5}~~\textbf{43.7} & 44.1~~\textbf{53.3}~~\makebox[\widthof{00.0}][c]{-} & 39.8~~49.4~~40.1 & 36.6~~44.7~~36.6                            & 42.4~~48.9~~\makebox[\widthof{00.0}][c]{-} & 42.4~~48.9~~\makebox[\widthof{00.0}][c]{-} \\
		length    & 15.7~~15.2~~15.8          & 16.4~~15.4~~16.9                            & 16.5~~15.4~~\makebox[\widthof{00.0}][c]{-}          & 24.8~~19.7~~26.9 & 27.0~~21.8~~29.0                            & 20.1~~19.0~~\makebox[\widthof{00.0}][c]{-} & 20.1~~19.0~~\makebox[\widthof{00.0}][c]{-} \\ \hline\hline
	\end{tabular}
	\caption{Retrieval scores averaged over domains for triplets \textit{test-0.4}, \textit{test-0.6}, \textit{test-de}. $\delta=0.1$}
	\label{tab:main-retrieval-avg-scores}
\end{table*}

\subsection{Interactions Between Retrieval and Translation \label{ssec:interactions}}


In this section, we look at the interactions between retrieval and translation and systematically vary the retrieval component for the three architectures of \textsection\ref{ssec:integrating-retrieval}. Notably, we compare two filters (NGM and BM25) during training and also contrast with a filter-free version in inference. We also vary the edit costs and the number of retrieved examples.
\fydone{pas clair, à reformuler}
\fydone{May be also scores for the copy}



\subsubsection{Architectures Comparison}

\paragraph{Neural fuzzy augmentation}

We observed, in preliminary results, that NFA is insensitive to the retrieval setting at training time. We only report results for a model trained on the baseline setting NGM+LED ($\tau=0.3$), then used in inference in a filter-free setting.

\begin{table}[H]
	\centering\small
	\begin{tabular}{|l|lll|}
		\hline
		ranker           & \textit{test-0.4} & \textit{test-0.6} & \textit{test-de} \\ \hline\hline
		NGM+LED 1-1      & 55.1              & 64.3              & -                \\
		NGM+LED 3-1      & 54.8              & 63.9              & -                \\
		NGM+LED 3-2      & 54.8              & 64.2              & -                \\
		NGM+LED 3-3      & 54.9/\tn{44.6}    & 64.3/\tn{45.7}    & 41.6             \\ \hline
		BM25             & 54.7/\tn{44.6}    & 64.2/\tn{45.6}    & -                \\
		BM25$_c$         & 54.7/\tn{44.6}    & 64.2/\tn{45.6}    & -                \\
		LED              & 54.9/\tn{44.7}    & 64.4/\tn{45.7}    & 41.7             \\
		$\delta$-LCS     & 54.8/\tn{44.6}    & 64.3/\tn{45.7}    & 41.9             \\
		$\delta$-LCS$_c$ & 54.8/\tn{44.6}    & 64.3/\tn{45.7}    & 41.8             \\ \hline
	\end{tabular}
	\caption{\label{tab:bleu-nfa} Average BLEU (/\tn{COMET ($\times$100)}) scores for en-fr (11 domains) and de-en (5 domains) using NFA models. $k_t$-$k_i$ in NGM+LED denotes a model trained with $k_t$ examples, while inference uses $k_i$; $_c$ denotes contrastive ranking.}
\end{table}

\fydone{for LED  and LCS what is the filter?}Results in Table~\ref{tab:bleu-nfa} are very consistent and hardly vary across domains (see Table~\ref{tab:bleu-nfa-per-domain} in Appendix~\ref{appendix:per-domain}) and language pairs. \fydone{pointer vers appendix} This is a first important result that somehow consolidates observations already performed for this model, which seems to be robust with respect to variations in the retrieval strategy.
\fydone{We need to see more clearly where there is a filter; also where there is always 3 retrieval, also include German results, COMET scores, link to full table of results and variance.}
\maxsmalltodo{COMET scores pas encore calculés (trop tard...)}

\paragraph{Edit-based techniques}

Regarding edit-based approaches, we train Multi-Levenshtein Transformer (\mlevt{3}) on the same dataset, comparing a set of retrieval settings and both NGM and BM25 filters.\fytodo{we miss some comet}
We report the following results in Table~\ref{tab:bleu-mlevt}:
\begin{itemize}
	\item The setting used in the original \mlevt{3} paper: NGM+LED ($\tau=0.3$) both at training and inference time.
	\item The best-performing train and inference setting pairs, as identifiable in Appendix~\ref{appendix:cross-inference}, for NGM and BM25 filters separately.
	\item We evaluate our best overall training pipeline (BM25+LED) on a filter-free setup with varying ED costs (using $\delta=0.1$).
\end{itemize}

\begin{table}[!ht]
	\centering\small
	\begin{tabular}{|l|lll|}
		\hline
		filter+ranker     & \textit{test-0.4}       & \textit{test-0.6}       & \textit{test-de} \\ \hline\hline
		NGM+LED           & 43.9/\tn{26.2}          & 56.0/\tn{47.8}          & 29.4             \\ \hline
		best NGM+ED pair  & 45.5/\tn{\bf{32.8}}     & 56.9/\tn{49.7}          & -                \\
		best BM25+ED pair & 45.7/\tn{31.0}          & 57.1/\tn{49.9}          & -                \\ \hline
		LED ~~(k=1)       & 44.9/\tn{29.0}          & 55.7/\tn{46.6}          & -                \\
		LED ~~(k=2)       & 45.2/\tn{29.1}          & 56.7/\tn{48.5}          & -                \\
		LED               & 45.6/\tn{31.3}          & 57.3/\tn{\bf{50.7}}     & 33.4             \\
		$\delta$-LCS      & 45.9/\tn{31.4}          & \textbf{57.4}/\tn{50.5} & \textbf{33.9}    \\
		$\delta$-LCS$_c$  & \textbf{46.0}/\tn{31.3} & 57.2/\tn{49.4}          & 33.7             \\ \hline
	\end{tabular}
	\caption{\label{tab:bleu-mlevt} Average BLEU (/\tn{COMET ($\times$100)}) scores across all 11 (en-fr) or 5 (de-en) domains using \mlevt{3}.}
\end{table}

We observe here a stronger impact of retrieval on the downstream scores, with a large gain over the baseline (for test-0.4 and en-de). $0.1$-LCS slightly outperforms LED in most conditions and metrics, with a very small difference between the contrastive and non-contrastive versions of the ranking.

\paragraph{In-context learning}

We evaluate BLOOM in $k$-shot translation for $k=1$ and $3$ with two rankers: ED and BM25.\footnote{We also consider a ``random'' ranking policy for contrast, which yields comparatively very poor results that are about 15 BLEU points below the others for $k=3$. This confirms the findings of \citet[Table~1, p.~235]{moslem-etal-2023-adaptive} on the benefits of TM-based retrieval.} The retrieval scope is always ``in-domain''. We do not use any filter to ensure the retrieval of exactly $k$ examples for each test instance.\footnote{In some situations however, we could not find 3 candidates with a score greater than $0$.}
\fydone{German comet?}
Results in Table~\ref{tab:pure-bloom-bleu} show again small differences between retrieval techniques, with a positive effect of contrastive ranking policy, which yields the best results. For this architecture, we also note that retrieving at least very good match (e.g.\ \textit{test-0.6}) does not necessarily imply very high BLEU scores, contrarily to NFA and \tmlevt{}.
\todo{Attention a l'allemand pour les données de test OpenSub, Aquis}
\begin{table}[!ht]
	\centering\small
	\begin{tabular}{|ll|ccc|}
		\hline
		ranker           & $k$-shot & \textit{test-0.4}  & \textit{test-0.6}  & \textit{test-de} \\ \hline\hline
		LED              & 1        & 45.8/\tn{30.5}     & 50.0/\tn{30.7}     & -                \\
		LED              & 3        & 47.8/\tn{35.3}     & 51.5/\tn{35.1}     & 33.3             \\
		LED$_c$          & 3        & \bf 48.3/\tn{37.5} & \bf 52.0/\tn{37.3} & -                \\
		$\delta$-LCS     & 3        & 48.1/\tn{36.0}     & 51.6/\tn{36.3}     & 33.7             \\
		$\delta$-LCS$_c$ & 3        & 48.2/\tn{36.4}     & 51.7/\tn{36.8}     & \bf 33.9         \\
		BM25             & 1        & 45.8/\tn{28.7}     & 49.6/\tn{27.1}     & -                \\
		BM25             & 3        & 48.1/\tn{35.0}     & 51.8/\tn{34.8}     & -                \\
		BM25$_c$         & 3        & 48.1/\tn{34.8}     & 51.4/\tn{35.2}     & -                \\ \hline
	\end{tabular}
	\caption{\label{tab:pure-bloom-bleu} Average BLEU/\tn{COMET ($\times$100)} scores averaged accross \textbf{7} en-fr domains (\textit{test-0.4/6}) and 5 de-en domains (\textit{test-de}) for ICL ($k$-shot) for several filter-free retrieval setups. $_c$ denotes contrast; $\delta=0.1$.}
\end{table}
\fydone{commenter par rapport à Vilar}

\subsection{Complementary Analyses}

\paragraph{What makes a good set of examples?} 

We use a linear model and try to predict BLEU scores using the retrieval metrics (coverage, relevance, and length) of \textsection\ref{ssec:retrieval-quality} for \mlevt{3} and ICL w.r.t.\ coverage, relevance, and length. First, with \mlevt{3}, the linear model has an average squared residuals of $0.2$ BLEU.
On the other hand, a constant model, which supposes that the model-agnostic metrics are independent of BLEU, has an error of $1.2$. Coverage, relevance, and length have respective coefficients of $0.13$, $0.09$, and $0.03$. This highlights the importance of coverage and relevance measures in the explanation of BLEU performance.
As for ICL, the constant model has the same average residual error ($0.28$ against $0.26$), with respective coefficients of $0.04$, $0.03$, and $0.00$. Thus, ICL seems more robust to changes in the retrieved examples.


\paragraph{Copying input tokens} The \textit{copy rate}, introduced in \textsection\ref{ssec:metrics} measures how much the translation model exploits slices from the examples to produce its output. Pipelines with high covering examples systematically imply a higher \textit{copy rate}. We find that \textit{copy rate} is correlated with higher BLEU scores for \mlevt{3} and ICL; in contrast NFA fails to produce higher BLEU scores with increasing \textit{copy rate}.

\begin{table}[!ht]
	\centering\small
	\begin{tabular}{|l|ccc|}
		\hline
		                                                    & \textit{test-0.4} & \textit{test-0.6} & \textit{test-de} \\ \hline\hline
		\rowcolor{LightGray}
		NFA                                                 &                   &                   &                  \\ \hline\hline
		BM25                                                & 47.0              & 62.1              & -                \\
		BM25$_c$                                            & 46.3              & 61.3              & -                \\
		LED                                                 & 43.4              & 60.8              & 38.2             \\
		$\delta$-LCS                                        & 46.7              & 62.8              & 41.0             \\
		$\delta$-LCS$_c$                                    & \bf 47.5          & \bf 63.8          & \bf 41.3         \\ \hline\hline
		\rowcolor{LightGray}
		\mlevt{3}                                           &                   &                   &                  \\ \hline\hline
		NGM+LED                                             & 37.4              & 56.3              & 30.4             \\ \hline
		best BM25+ED pair                                   & 42.4              & 57.8              & -                \\ \hline
		LED                                                 & 40.7              & 56.9              & 37.2             \\
		$\delta$-LCS                                        & 42.8              & 58.3              & \bf 39.5         \\
		$\delta$-LCS$_c$                                    & \bf 43.0          & \bf 58.4          & 39.4             \\ \hline\hline
		\rowcolor{LightGray}
		\makebox[\widthof{xxxxxxxx}][l]{ICL} k-shot         &                   &                   &                  \\ \hline\hline
		\makebox[\widthof{xxxxxxxx}][l]{LED} 1              & 40.5              & 53.5              & -                \\
		\makebox[\widthof{xxxxxxxx}][l]{LED} 3              & 51.8              & 64.7              & 45.0             \\
		\makebox[\widthof{xxxxxxxx}][l]{LED$_c$} 3          & 53.7              & 65.9              & -                \\
		\makebox[\widthof{xxxxxxxx}][l]{$\delta$-LCS} 3     & 54.1              & 66.1              & 47.2             \\
		\makebox[\widthof{xxxxxxxx}][l]{$\delta$-LCS$_c$} 3 & 54.3              & 65.2              & \bf 48.1         \\
		\makebox[\widthof{xxxxxxxx}][l]{BM25} 1             & 53.9              & 66.1              & -                \\
		\makebox[\widthof{xxxxxxxx}][l]{BM25} 3             & 54.1              & 66.1              & -                \\
		\makebox[\widthof{xxxxxxxx}][l]{BM25$_c$} 3         & \bf 54.9          & \bf 67.0          & -                \\ \hline
	\end{tabular}
	\caption{Average \textit{copy rate} of all three models.}
	\label{tab:copy-rate-bloom}
\end{table}

Also note that, even though it relies on explicit edits, \mlevt{3} always has a lower \textit{copy rate} than NFA or ICL; the latter notably has the highest \textit{copy rate} for \textit{test-0.6} and also the worst translation scores, suggesting that too many irrelevant tokens are kept in the output.
\fydone{Est-ce que c'est pertinent pour notre propos ?}

\paragraph{Domain Filtering}
Relaxing the constraint that similar examples should be retrieved ``in-domain'' increases the retrieval rate. However, it turns out to be detrimental for all architectures: results are in Table~\ref{tab:domain-mlevt}, where we compare in-domain retrieval with all-domains and out-of-domain retrievals.

\begin{table}[!ht]
	\centering\small
	\begin{tabular}{|l|ccc|}
		\hline
		domain & NFA         & \mlevt{3}   & ICL         \\ \hline\hline
		In     & 54.9 / 64.4 & 45.6 / 57.3 & 47.8 / 51.5 \\
		All    & 52.5 / 62.3 & 43.8 / 55.4 & 45.1 / 49.0 \\
		Out    & 45.5 / 51.2 & 30.2 / 33.5 & 30.7 / 29.6 \\ \hline
	\end{tabular}
	\caption{\label{tab:domain-mlevt} Average BLEU scores (\textit{test-0.4} / \textit{test-0.6}, en-fr) according to the domain selection strategy, using filter-free LED as ranker. }
\end{table}

The impact of the ``all-domain'' policy is genuine: when this policy is enforced, the most similar examples are found out-of-domain for 35.1\% of our 22k test samples. The most impacted domains are Ubuntu (82.8\% matches are out-of-domain) and NewsCommentary (79.7\%). The per-domain analysis (Appendix~\ref{appendix:per-domain}) however shows that this is detrimental for Ubuntu (-6.1/-6.7 BLEU for \textit{test-0.4}/\textit{test-0.6}), and neutral for NewsCommentary (-0.2/+0.2 BLEU). As expected, enforcing an ``out-of-domain'' selection constraint yields dramatic losses in BLEU (-15.4/-23.8 BLEU).

These results confirm the benefit of retrieving ``in-domain'', even for small domains: not only does it greatly speed up retrieval, but it also yields better examples and, ultimately, higher translation scores.

\paragraph{Simplifying the Retrieval Pipeline}
For large domains, removing the filtering step in the retrieval pipeline considerably increases the computational cost,
especially during training.\footnote{The complexity is quadratic w.r.t.\ to the TM size.} Yet, it may prematurely discard useful examples, especially when using contrastive ranking. To evaluate this, we turn off filtering for test samples during inference.
This simplification of the pipeline improves the scores for \mlevt{3}, while there is no effect for NFA (see lines for filtering free inference in Tables~\ref{tab:bleu-nfa} and \ref{tab:bleu-mlevt}). Thus, for the former method at least, a trade-off can be made between latency and translation scores.

\paragraph{Increasing the number of examples} We vary $k$, the number of TM examples retrieved, from $1$ to $3$. Overall, we observe a gain (BLEU/COMET) when $k$ increases. For ICL, this is already clear from the results in Table~\ref{tab:pure-bloom-bleu} where 3-shots clearly outperforms 1-shot.
\fytodo{No comet score, no increase for NFA}
We get a similar conclusion for \mlevt{3} based on the results in Table~\ref{tab:bleu-mlevt}, where we vary the inference procedure for a model trained on (BM25+LED). The test retrieval is filter-free LED with either exactly $k=1$, $2$, or $3$ retrieved examples.

As for NFA, a model trained on up to $3$ instances slightly benefits from more examples but does not compete with a model using only the one-best match in training and inference. This seems to contradict \citet{bulte-tezcan-2019-neural}, who claim the superiority of using more examples.


\paragraph{Optimizing for coverage with $\delta$-LCS}

By design, $\delta$-LCS retrieves examples having a higher coverage of the source than LED, which turns into higher \textit{copy rates} for all architectures.
For ICL and \mlevt{3}, it yields similar BLEU gain (ICL: +0.3/+0.1/+0.4; \mlevt{3}: +0.3/+0.1/+0.5 on resp. \textit{test-0.4}, \textit{test-0.6}, \textit{test-de}). The analysis in Appendix~\ref{appendix:cross-inference} shows a consistent benefit of $\delta$-LCS for \mlevt{3} when coupled with filters at training (NGM) and inference time (NGM and BM25).
\fydone{So ?}

\paragraph{Enforcing diversity with contrastive ranking}
We observe that using a contrastive ranker is mostly beneficial for medium-scoring similar examples (\textit{test-0.4}), regardless of the architecture. It can even be detrimental when at least one high-matching example is found. This is because contrastive ranking generates less similar examples that are not necessarily relevant. In comparison, for \textit{test-0.4}, increasing the diversity in retrieval seems beneficial, as it increases the coverage of the source. This suggests that contrastive methods should adapt their strength parameter ($\alpha$ in \eqref{eq:contrast}) to the retrieval scores, enforcing more diversity when matches are poor.

\section{Related Work \label{sec:related}}

\fydone{copied from MLEVT - to be revised and updated with recent refs}
As for other text generation applications \cite{li2022survey}, efforts to integrate a retrieval component in NMT have intensified in recent years. One motivation is to increase the transparency of ML models by providing users with the retrieved examples that were used to compute the actual output \citep{rudin-cynthia-2019-stop}. For MT, this is achieved by integrating fuzzy matches retrieved from memory as an additional context. This can be performed by concatenating the retrieved target instance to the source text, an approach that also accommodates several TM matches \cite{bulte-tezcan-2019-neural}, or by the simultaneous use of their source and target sides \cite{pham-etal-2020-priming,reheman-etal-2023-prompting}. More complex schemes to combine retrieved examples with the source sentence are in \citep{gu-etal-2018-search,xia-etal-2019-graph,he-etal-2021-fast}. 
The recent studies of \citet{cheng-etal-2022-neural,agrawal-etal-2023-context} and \citet{sia-and-duh-2023-incontext} handle several complementary TM examples retrieved in a \emph{contrastive manner} that aims to enhance source coverage.
\citet{gupta-etal-2023-coverage} propose a general formulation in the application of ICL in various tasks.
\citet{cai-etal-2021-neural} also handle multiple matches and introduce two novelties: (a) retrieval is performed directly in the target language and (b) similarity scores are trainable, which allows to evaluate retrieved instances based on their usefulness in translation. Most of these attempts rely on auto-regressive (AR) decoding, meaning that the impact of TM match(es) on the output is only indirect.

The use of TM memory match with a NAT decoder is studied by \citet{niwa-etal-2022-nearest,xu-etal-2023-integrating,zheng-etal-2023-towards}, who adapt \levt{} for this specific setting, using one single retrieved instance to initialize the edit-based decoder; \cite{bouthors-etal-2023-towards} extends this technique to process multiple retrieved examples.
\citet{zhang-etal-2018-guiding} explore a different set of techniques to improve translation using retrieved segments instead of full sentences. Generalizing nearest neighbor language models (NNLMs) \citep{he-etal-2021-efficient} to conditional LMs, \citet{khandelwal_nearest_2021} perform $k$-NNMT as follows: at each decoding step, the $k$ target contexts that closest to the current contextualized representations are retrieved and used to select the next token. This approach is further elaborated in \cite{zheng-etal-2021-adaptive,meng-etal-2022-fast} and extended to chunks by \citet{martins-etal-2022-chunkbased}. \fytodo{Check more recent papers ?}


A final thread of relevant papers concerns the use of large language models, which, provided with suitable prompts and in-context examples, can be turned into effective translation systems. Such approaches have been tested with most LLMs,
with the goal to illustrate the multi-tasking abilities of such models. Closer to our work, a series of work have tried to optimize LLMs performance for the MT task, systematically studying the effect of the prompt change, of the number of shots, and of the in-context examples selection procedures \citep{vilar-etal-2023-prompting,zhang-et-al-2023-prompting,hendy-etal-2023-howgood,bawden-yvon-2023-investigating}. \citet{moslem-etal-2023-adaptive} were the first to combine LLMs with TMs, using an embedding-based retrieval system and combining (via concatenation) up to 5 TM-matches in the MT prompt;
\cite{mu-etal-2023-augmenting} followed suit, with a different LLM and a two-stage retrieval strategy (first 500 closest matches for a Lucene-based engine; then using up to 9 closest matches for the edit-distance).
\cite{agrawal-etal-2023-context} studies a way to optimally select $k$ examples so as to maximize coverage, an approach akin to our "contrastive'' scenario -- using a BM25 retriever in a first stage, and a greedy heuristic selection in a second stage. \cite{sia-and-duh-2023-incontext} explores another benefit of selecting good in-context examples, that of maintaining consistency in the generated text - for this, they retrieve examples from a moving context window of past translations.\fytodo{Note that they use TED and small BLOOM}
Finally,  \citet{m-etal-2023-ctqscorer} go one step further by training a linear regression model predicting the goodness of TM instances based on a small set of features.

\section{Conclusion \label{sec:conclusions}}
This paper has investigated the effect of varying the retrieval strategy for three commonly used retrieval-augmented machine translation architectures, trying to get a better understanding of the interplay of these two components. While auto-regressive encoder-decoder architecture seems quite robust w.r.t.\ changes in the retrieval strategy, this is less so for the two other architectures, for which optimizing the retrieval policy can yield significant returns. Our experiments have also highlighted the benefits of coverage-oriented retrieval policies, based on LCS, especially for the non-autoregressive model. Finally, we have validated the use of the ``in-domain'' selection policy and proposed to simplify the inference step by eliminating the filtering process, yielding better performance at the expense of an increased latency.

In our future work, we would like to continue the exploration of the interplay between retrieval and translation, with the aim to jointly optimize these two processes rather than have them designed independently. \fytodo{a note about the difference between BM25 and ED, and the fact that the former is cheaper and easier to optimize}
\fytodo{ajouter remarque sur qualité exemples src-tgt}

\section{Limitations \label{sec:limitations}}

In this paper, we have focused on purely transductive techniques, meaning that all inference is performed with a frozen network - this limitation certainly needs to be reconsidered, and our results would be stronger with additional comparisons with e.g., on-the-fly fine-tuning \citep{farajian-etal-2017-multi} or low-rank adaptation techniques \cite{hu2022lora}.

We have chosen to use only one large LLM, with 176b million parameters. This was motivated by (a) the openness of the model and the transparency of the training data, which allowed us to control for test samples occurring also in the training; (b) the existence of multiple previous experiments with this model, which allowed us to get a reasonable idea of its basic translation abilities. More recent, smaller, and arguably better models (e.g., the LLAMA \cite{touvron2023llama} and Falcon families) \cite{almazrouei2023falcon} with various levels of multilingual support \citep{alves2024tower}, would likely yield a more faithful picture of the current performance of in-context learning with LLMs.

Our discussion has focused on measures of translation quality; in practical applications, computational costs associated with a specific combination of retrieval and architecture also matter. While we have tried to be explicit about the complexity of each retrieval algorithm, we have left aside issues related to identifying the optimal computation/performance tradeoffs.

\section{Acknowledgements}

This work was performed using HPC/AI resources from GENCI-IDRIS (Grants 2022-AD011013583 and 2023-AD010614012). The third author was also partly funded by the French National Research Agency (ANR) under grant ANR-23-IAS1-0006 (TraLaLam).

\bibliography{anthology,custom}



\appendix

\section{$\delta$-LCS formulation}
\label{appendix:d-lcs}

The general formulation of a similarity metric based on the edit distance is:
\begin{equation}
	\simi (\tilde \src, \src) = 1 - \frac{\Delta(\tilde \src, \src)}{N (|\tilde \src|, |\src|)},
	\label{eq:ed}
\end{equation}
with $N$ a normalizing term that upper bounds $\Delta$. $\Delta$
is parameterized by $d$, $a$, $r$, corresponding respectively to the delete (resp. insert, replace) costs (assuming copies have a $0$ cost), generalizing the standard ED setting ($d=a=r=1$). $\Delta$ corresponds to the minimal generalized cost necessary to edit $\tilde \src$ into $\src$.

We define an Edit Common Subsequence $\ecs_{d,a,r}(\tilde \src, \src)$, which corresponds to the tokens copied when optimally editing $\tilde \src$ into $\src$.\footnote{There can be in fact several concurrent common subsequences.} If $a + d \leq r$, then the copied tokens are a Longest Common Subsequence (LCS) of $\tilde \src$ and $\src$: $\ecs_{d,a,r}(\tilde \src, \src) = \lcs(\tilde \src, \src)$.

If $a + d > r$:
\begin{equation}
	\resizebox{.95\hsize}{!}{$
			\Delta(\tilde \src, \src) = \left\{
			\begin{array}{l}
				r \left[ |\src| -         |ecs_{d,a,r}(\tilde \src, \src)| \right] + d (|\tilde \src| - |\src|) \\
				~~~~~\text{if}~~ |\src| \leq |\tilde \src|                                                      \\
				r \left[ |\tilde \src| -  |ecs_{d,a,r}(\tilde \src, \src)| \right] + a (|\src| - |\tilde \src|) \\
				~~~~~\text{if}~~ |\src| > |\tilde \src|                                                         \\
			\end{array}
			\right.
		$}
\end{equation}

If $a + d \leq r$:
\begin{equation}
	\resizebox{.95\hsize}{!}{$
			\Delta(\tilde \src, \src) = a (|\src| - |\lcs(\tilde \src, \src)|) + d (|\tilde \src| - |\lcs(\tilde \src, \src)|)
		$}
\end{equation}

$\Delta$ is maximal when $|\ecs| = 0$.
The normalization term can be expressed as:
\begin{equation}
	\resizebox{.95\hsize}{!}{$
			N_{d, a, r}(|\tilde \src|, |\src|) = \left\{
			\begin{array}{l}
				a |\src| + d |\tilde \src|                                               \\
				~~~~~~~~~~\text{if}~~ a + d \leq r                                       \\
				(r - d) |\src| + d |\tilde \src|                                         \\
				~~~~~~~~~~\text{if}~~ a + d > r ~~\text{and}~~ |\src| \leq |\tilde \src| \\
				(r - a) |\tilde \src| + a |\src|                                         \\
				~~~~~~~~~~\text{if}~~ a + d > r ~~\text{and}~~ |\src| > |\tilde \src|
			\end{array}
			\right.
		$}
\end{equation}

Choosing $(d, a, r) = (0, 1, 1)$ makes $\simi (\tilde \src, \src)$ compute the coverage of $\src$ with tokens from $\tilde \src$:
\begin{equation}
	\resizebox{.8\hsize}{!}{$
			\simi(\tilde \src, \src) = 1 - \frac{a(|\src| - |\lcs(\tilde \src, \src)|)}{a |\src|} = \frac{|\lcs(\tilde \src, \src)|}{|\src|}
		$}
\end{equation}

\section{Fuzzy-Matching detailed settings}
\label{appendix:fm-settings}

Note that the code used for our experiments is open source.\footnote{Available at
	\href{https://github.com/SYSTRAN/fuzzy-match}{https://github.com/SYSTRAN/fuzzy-match}
} Additional details can be found in the repository.

\subsection{NGM}
\label{appendix:ngm}

For a source $q$, n-gram matching identifies $g(q, x)$, the longest common n-gram between $q$ and any source example $x$. $x$ passes the filter if (1) $|g(q, x)| \geq \text{ML}$, an absolute length threshold that we choose to be $3$; (2) $|g(q, x)| \geq \tau |q|$, with $\tau$ varying between $0.3$ (following the works of \citet{xu-etal-2022-boosting,bouthors-etal-2023-towards}) and $0.2$ -- which is a more permissive filter that increases the chance to having higher scoring matches w.r.t.\ the ranker (BM25, ED).

Our algorithm uses a suffix array structure which directly indexes the n-grams and enables to search for n-gram matches in the sorted array with a logarithmic complexity.

However, the distribution of the number of candidates passing the filter has a high variance (see Table~\ref{tab:quartiles-ngm}). In consequence, it can be ineffective to select only a small amount of relevant candidates (too permissive). On the other hand, it can often retrieve very few candidates (<3).

\begin{table}[H]
	\centering
	\begin{tabular}{|l|rrr|}
		\hline
		$\tau$ & Q1 & Q2  & Q3   \\ \hline\hline
		0.2    & 6  & 116 & 1573 \\
		0.3    & 2  & 15  & 293  \\
		0.4    & 1  & 6   & 99   \\
		0.5    & 1  & 4   & 54   \\ \hline
	\end{tabular}
	\caption{\label{tab:quartiles-ngm} Average 1st, 2nd, 3rd quartiles for the eleven datasets of the distribution of candidates passing the NGM filter w.r.t.\ $\tau$.}
\end{table}

\subsection{BM25}
\label{appendix:bm25}

BM25 \citep{robertson2009probabilistic} is a widely used ranker in Information Retrieval. It often competes with state-of-the-art neural retrievers and is used as a baseline. It is a fully unsupervised method based on computations inspired by the \emph{tf-idf} scoring function. BM25 typically aims to retrieve documents that have common rare words with the query.

Our implementation is not as optimized as in Elasticsearch\footnote{\url{https://elastic.co}.} or Apache Lucene\footnote{\url{http:lucene.apache.org}.}, \fydone{references} it is nonetheless very efficient and integrates seamlessly in our translation pipeline.\fytodo{Can we have numbers?}

Since computing BM25 is computationally linear in the number of sentences in the TM, naive implementations can sometimes be slow. We use an \emph{inverted index}, which directly maps each term to the set of TM segments where they occur. Instead of computing BM25 for every sentence, we only do it for the union of TM segments containing query terms. As common terms such as punctuation, prepositions, etc. will likely map to most TMs, they are removed from the index.

In our implementation, a term is said to be common if it appears in more than $p\%$ of the segments. We find that $p$ can be quite small (2\%) without affecting the set of retrieved sentences too much. For instance, we find that BM25+LCS with the setting $L=100$, $p=2\%$ has $90.4\%$ Jaccard similarity (in terms of the set of indices of retrieved segments) with the setting $L=100$, $p=10\%$.

In preliminary experiments, we also varied the value of $L$ ($10$ vs $100$) and found out that a low $L$ negatively affects the model-agnostic metrics (coverage, relevance) and set $L=100$ in all experiments.\fytodo{check this}

Table~\ref{tab:appendix-retrieval-avg-scores} contains the model-agnostic scores on the training set en-fr for filtering pipelines. It suggests that BM25 filter is better at increasing coverage and relevance than NGM. But it has a higher latency. This can explain why BM25-filtered pipelines are better at training time (as shown in Appendix~\ref{appendix:cross-inference}).

\section{Models detailed settings}
\label{appendix:models-settings}

\subsection{NFA}
\label{ssec:nfa-settings}

We made our own implementation of NFA based on the work of \citet{bulte-tezcan-2019-neural} with the following parameters: size of word embedding: $512$; size of hidden layers: $512$; size of inner feed forward layer: $2,048$; number of heads: $8$; number of layers: $6$; batch size: $4,096$ tokens.
We set warmup steps to $4,000$ and update learning rate for every $8$ iterations.
Fine-tuning is performed continuing Adam with learning rate decay schedule until convergence. The models are trained with one NVIDIA P100 GPU. We use a joint vocabulary of $32$K for both source and target sides. At inference we use a beam size of $5$.

\subsection{\mlevt{N}}
\label{ssec:mlevt-settings}

We use a Multi-Levenshtein Transformer architecture with embeddings of dimension $512$; feed-forward layers of size $2048$; number of heads $8$; number of encoder and decoder layers: 6; shared-embeddings; dropout: 0.3.

During training, we use Adam optimizer with ($\beta_1$, $\beta_2$)=(0.9, 0.98); inverse sqrt scheduler; learning rate: $5e^{-4}$; label smoothing: 0.1; warmup updates: 10,000; float precision: 16. We fixed the number of iterations at $60$k. The batch size and number of GPUs are set to have, on average, $\sim 450$ samples per iteration. We performed a pretraining of the model on a synthetic dataset as described by \citet{bouthors-etal-2023-towards}.
We use a joint vocabulary of $32$K.
For decoding, we use realignment and iterative refinement with an empty placeholder penalty of $3$, and a max number of iterations of $10$ \cite{gu-etal-2019-levenshtein}.

\subsection{BLOOM \label{ssec:bloom}}
BLOOM is a family of large open-source causal multilingual language models trained in the course of the BigScience project \cite{lescao2022bloom}. Our experiments use the largest available version, comprising 176b parameters. This model has been repeatedly evaluated in translation scenarios see e.g.\ \cite{bawden-yvon-2023-investigating}. BLOOM's training corpus officially contains a fair share of English and French, but hardly any German\footnote{Even though traces of German can be found in the corpus.} \cite{laurencon2022bigscience}. We thus consider the two following translation directions: en-fr and de-en, selecting target languages for which the generation abilities are strong.

Following common practice, we select a simple prompt of the form ``[src]: source sentence. =[trg]: target sentence'', where ``[src]'' and ``[trg]'' are language tags denoting respectively the source and the target sentences, using either 1 or 3 in-context examples before the source query. All experiments generate translation in pure greedy mode; generation stops either when the end of sentence token is produced or when a maximum length of 256 tokens is reached. As is also custom for BLOOM, which tends to overgenerate, we post-process the output to shorten excessively long outputs and make sure the target is never longer than 1.5 times the source (measured in chars). This post-processing only impacts about 5 to 10\% of all output sentences. \fydone{check this}

\begin{table*}
	\centering\small
	\begin{tabular}{|l|*{9}{c|}} \hline
		Ngram filter ($\tau=$)           & $0.3$ & $0.2$ & $0.2$        & $0.3$        & $0.2$        & -         & -         & -            & -            \\
		BM25 filter ($L=$ or '-')        & -     & -     & -            & -            & -            & 100       & 100       & 100          & 100          \\
		Ranking (LED, LCS, $\delta$-LCS) & LED   & LCS   & $\delta$-LCS & $\delta$-LCS & $\delta$-LCS & LED       & LCS       & $\delta$-LCS & $\delta$-LCS \\
		Contrast factor ($\alpha=$)      & -     & -     & -            & -            & $0.3$        & -         & -         & -            & $0.3$        \\ \hline\hline
		                                 &       &       &              &              &              &           &           &              &              \\[-1.9ex]
		coverage                         & 27.9  & 42.5  & 38.8         & 35.0         & 39.5         & 32.6      & \bf{48.1} & 43.6         & 44.6         \\
		relevance                        & 23.0  & 23.3  & 26.1         & 23.9         & 24.6         & 26.6      & 28.8      & \bf{29.2}    & 27.6         \\
		length                           & 15.4  & 40.6  & 24.0         & 22.0         & 25.5         & \bf{15.2} & 30.1      & 22.4         & 23.5         \\ \hline\hline
	\end{tabular}
	\caption{Retrieval scores averaged over 11~domains (train sets, en-fr).}
	\label{tab:appendix-retrieval-avg-scores}
\end{table*}

\section{Cross inference for \mlevt{3}}
\label{appendix:cross-inference}

\begin{table*}[!ht]
	\centering\small
	\begin{tabular}{|l|rrrrr||rrrrr|}
		\hline
		train $\backslash$ infer & LED$^{+}$        & LCS$^{-}$        & $\delta$-LCS$^{-}$        & $\delta$-LCS$^{+}$ & $\delta$-LCS$^{-}_c$      & LED$^{+}$        & LCS$^{-}$        & $\delta$-LCS$^{-}$        & $\delta$-LCS$^{+}$        & $\delta$-LCS$^{-}_c$ \\ \hline\hline
		LED$^{+}$                & 43.9             & 44.1             & \textbf{44.8}             & \textbf{44.8}      & \textbf{44.8}             & 56.0             & 55.6             & 56.2                      & \textbf{56.3}             & 56.1                 \\
		LCS$^{-}$                & 43.9             & 44.5             & 44.8                      & 44.6               & \textbf{45.0}             & 55.7             & 56.2             & 56.5                      & 56.3                      & \textbf{56.6}        \\
		$\delta$-LCS$^{-}$       & \underline{44.6} & \underline{44.8} & \underline{\textbf{45.3}} & \underline{45.2}   & \underline{\textbf{45.5}} & 55.9             & 56.1             & \textbf{56.6}             & 56.5                      & 56.5                 \\
		$\delta$-LCS$^{+}$       & 44.1             & 44.5             & \underline{45.3}          & 45.1               & \textbf{45.3}             & \underline{56.3} & \underline{56.3} & \underline{\textbf{56.9}} & \underline{\textbf{56.9}} & \underline{56.8}     \\
		$\delta$-LCS$^{-}_c$     & 43.8             & 44.1             & 44.7                      & 44.5               & \textbf{44.9}             & 55.6             & 55.9             & \textbf{56.4}             & 56.2                      & 56.3                 \\ \hline
	\end{tabular}
	\caption{\label{tab:cross-inference-mlevt-ngm} Average cross-inference BLEU score accross all domains for \textit{test-0.4} (left) and \textit{test-0.6} (right) with NGM filter. $\tau$ is specified with $^+$ for $0.3$ and $^-$ for $0.2$; $c$ denotes contrast. Best column-wise (resp. row-wise) BLEU are \underline{underlined} (resp. \textbf{in bold}).}
\end{table*}

\begin{table*}[!ht]
	\centering\small
	\begin{tabular}{|l|rrrr||rrrr|}
		\hline
		train $\backslash$ infer & LED              & LCS              & $\delta$-LCS              & $\delta$-LCS$_c$          & LED              & LCS              & $\delta$-LCS              & $\delta$-LCS$_c$ \\ \hline\hline
		LED                      & \underline{45.1} & \underline{45.2} & \textbf{\underline{45.7}} & \textbf{\underline{45.7}} & \underline{57.0} & \underline{56.5} & \textbf{\underline{57.1}} & \underline{57.0} \\
		LCS                      & 44.7             & 45.1             & 45.3                      & \textbf{45.4}             & 56.1             & 56.4             & \textbf{56.5}             & 56.4             \\
		$\delta$-LCS             & 44.6             & 45.1             & 45.3                      & \textbf{45.4}             & 56.4             & 56.6             & 56.8                      & \textbf{56.9}    \\
		$\delta$-LCS$_c$         & 42.3             & 43.2             & 43.6                      & \textbf{44.0}             & 55.2             & 55.6             & 55.8                      & \textbf{56.1}    \\ \hline
	\end{tabular}
	\caption{\label{tab:cross-inference-mlevt-bm25} Average cross-inference BLEU score accross all domains for \textit{test-0.4} (left) and \textit{test-0.6} (right) with BM25 filter. $c$ designates contrast. Best column-wise (resp. row-wise) BLEU are \underline{underlined} (resp. \textbf{in bold}).}
\end{table*}

We train Multi-Levenshtein Transformer (\mlevt{3}) with a series of retrieval settings with NGM (Table~\ref{tab:cross-inference-mlevt-ngm}) and BM25 filters (Table~\ref{tab:cross-inference-mlevt-bm25}).
Afterward, each trained model is evaluated on all the retrieval settings of the same filter category.
This way, it is possible to identify which train/inference setting pairs perform better and whether having the same setting for training and inference is recommended.
We can, for both NGM and BM25, identify a systematically independent optimal training and inference setting. We find that the BM25 filter is slightly better in general. The contrastive method only helps at inference time and for medium matches, suggesting adapting the factor to the fuzzy score (ED).

$0.1$-LCS is overall better, except for training with BM25, which surprisingly achieves the best scores with LED. Probably, having more source-similar sentences at train time is necessary to compensate for the unordered nature of BM25, required by \mlevt{3}.

\section{Per domain analysis}
\label{appendix:per-domain}

Tables~\ref{tab:bleu-nfa-per-domain},\ref{tab:bleu-mlevt-per-domain},\ref{tab:bleu-bloom-per-domain} contain the detailed per-domain analysis for direction en-fr.
We insist on the fact that this corpus offers a high variability across domains (size, density, sentence length...).
Low-density domains (Ubuntu, News-Commentary) and high-density ones (EMEA, KDE4) are the ones the most likely to differ from the average best.
Moreover, low-density domains seem to prefer diversity and coverage with BLOOM. As for $\mlevt{3}$, the lowest and highest densities are both more inclined to choose pipelines with filters. The model may suffer from low-quality examples from a difficult low-density domain while not necessitating diversity in the case of dense ones.

\begin{table*}[!ht]
	\centering
	\small
	\begin{tabular}{l|rrrrrrrrrrr|r}
		\hline
		\rowcolor{LightGray}
		pipeline             & ECB      & EME      & Epp      & GNO      & JRC      & KDE      & News     & PHP      & TED      & Ubu      & Wiki     & \textbf{all} \\ \hline\hline
		~~~\textit{test-0.4} &          &          &          &          &          &          &          &          &          &          &          &              \\
		NGM+LED 1-1          & 64.7     & 65.3     & 44.9     & \bf 71.2 & 76.4     & \bf 65.9 & \bf 30.4 & 42.0     & 43.9     & 57.5     & 43.5     & \bf 55.1     \\
		NGM+LED 3-3          & 64.8     & 65.8     & 44.6     & 71.1     & 76.4     & 65.4     & 29.7     & 41.4     & \bf 44.0 & 56.4     & 44.4     & 54.9         \\
		NGM+LED 3-2          & 64.9     & 65.6     & 44.5     & 70.8     & 76.3     & 65.4     & 29.7     & 41.4     & \bf 44.0 & 56.3     & 44.2     & 54.8         \\
		NGM+LED 3-1          & 64.4     & 65.0     & 44.7     & 70.6     & 76.2     & 65.7     & 29.8     & 41.6     & 43.9     & 56.6     & 44.1     & 54.8         \\
		BM25                 & 64.1     & 66.3     & 44.8     & 70.8     & 76.0     & 65.0     & 30.2     & 40.0     & 43.5     & \bf 57.7 & 43.0     & 54.7         \\
		BM25$_c$             & 64.2     & \bf 66.4 & 44.8     & 70.9     & 76.0     & 64.9     & 30.2     & 40.0     & 43.5     & 57.6     & 43.0     & 54.7         \\
		LED                  & 64.8     & 65.4     & 44.8     & 71.1     & \bf 76.5 & 65.4     & 29.6     & 41.0     & 43.8     & 56.7     & \bf 44.5 & 54.9         \\
		$\delta$-LCS         & 65.0     & 64.6     & 44.8     & 70.9     & \bf 76.5 & 65.3     & 29.9     & 41.4     & 43.6     & 57.3     & 43.9     & 54.8         \\
		$\delta$-LCS$_c$     & \bf 65.2 & 63.9     & \bf 45.0 & 70.8     & 76.4     & 65.4     & 29.8     & \bf 41.8 & 43.9     & 57.3     & 43.6     & 54.8         \\ \hline\hline
		~~~\textit{test-0.6} &          &          &          &          &          &          &          &          &          &          &          &              \\
		NGM+LED 1-1          & 71.0     & \bf 73.3 & 59.1     & 79.9     & 83.7     & 68.1     & \bf 27.6 & \bf 46.4 & 63.7     & 64.4     & 69.8     & 64.3         \\
		NGM+LED 3-3          & 70.8     & 73.1     & 59.1     & 80.4     & \bf 83.9 & \bf 69.4 & 27.3     & 45.7     & 64.2     & 64.1     & 69.7     & 64.3         \\
		NGM+LED 3-2          & 70.8     & 72.3     & 59.1     & 80.4     & 83.7     & 69.2     & 27.2     & 45.6     & 64.1     & 63.8     & \bf 70.1 & 64.2         \\
		NGM+LED 3-1          & 70.1     & 72.2     & 59.0     & 79.8     & 83.6     & 68.5     & 27.4     & 44.8     & 64.1     & 63.5     & 69.5     & 63.9         \\
		BM25                 & \bf 71.2 & 72.8     & \bf 59.2 & 81.0     & 83.5     & 67.7     & 27.5     & 44.8     & \bf 64.3 & \bf 64.9 & 69.1     & 64.2         \\
		BM25$_c$             & \bf 71.2 & 72.9     & \bf 59.2 & \bf 81.1 & 83.5     & 67.7     & \bf 27.6 & 44.8     & \bf 64.3 & 64.8     & 68.9     & 64.2         \\
		LED                  & 70.8     & 73.1     & 59.1     & 80.4     & 83.8     & \bf 69.4 & 27.3     & 45.7     & 64.1     & 64.2     & 69.9     & \bf 64.4     \\
		$\delta$-LCS         & 71.1     & 73.2     & 59.1     & 80.1     & 83.7     & 68.7     & 27.2     & 46.0     & 64.2     & 64.4     & 69.7     & 64.3         \\
		$\delta$-LCS$_c$     & 71.1     & 73.0     & 59.1     & 80.2     & 83.8     & 68.9     & 27.1     & 45.9     & \bf 64.3 & 64.4     & 69.5     & 64.3         \\
	\end{tabular}
	\caption{BLEU score (en-fr): NFA trained on NGM+LED ($\tau=0.3$), inferred on filter-free retrieval settings.}
	\label{tab:bleu-nfa-per-domain}
\end{table*}

\begin{table*}
	\centering
	\small
	\begin{tabular}{l|rrrrrrrrrrr|r}
		\hline
		\rowcolor{LightGray}
		pipeline             & ECB      & EME      & Epp      & GNO      & JRC      & KDE      & News     & PHP      & TED      & Ubu      & Wiki     & \textbf{all} \\ \hline\hline
		~~~\textit{test-0.4} &          &          &          &          &          &          &          &          &          &          &          &              \\
		best NGM+ED pair     & 57.4     & \bf 57.6 & 35.8     & 62.0     & 68.7     & 55.9     & \bf 22.1 & 29.2     & 31.6     & 45.2     & \bf 35.8 & 45.6         \\
		best BM25+ED pair    & 57.5     & 57.0     & 35.8     & 63.0     & 69.2     & \bf 57.5 & 21.9     & 30.4     & 31.3     & 45.5     & 34.8     & 45.8         \\ \hdashline
		LED                  & 56.7     & 56.9     & 35.6     & 62.4     & 68.8     & 56.4     & 21.5     & 30.6     & 32.3     & 45.9     & 34.4     & 45.6         \\
		$\delta$-LCS         & 57.3     & 56.5     & \bf 36.8 & \bf 63.3 & 69.2     & 56.7     & 21.9     & \bf 30.9 & \bf 31.9 & 45.1     & 34.9     & 45.9         \\
		$\delta$-LCS$_c$     & \bf 57.7 & 56.8     & 36.4     & 62.8     & \bf 69.3 & 56.8     & 21.8     & 30.8     & 31.7     & \bf 46.5 & 35.2     & \bf 46.0     \\ \hline\hline
		~~~\textit{test-0.6} &          &          &          &          &          &          &          &          &          &          &          &              \\
		best NGM+ED pair     & 65.8     & \bf 68.3 & 51.6     & 73.3     & 77.5     & \bf 63.0 & 21.2     & 33.4     & 55.9     & 53.0     & 63.6     & 57.0         \\
		best BM25+ED pair    & 66.7     & 67.8     & 51.2     & 72.6     & 78.4     & 62.5     & \bf 21.4 & 34.9     & 55.9     & 53.7     & 63.6     & 57.2         \\ \hdashline
		LED                  & 66.2     & 68.1     & 51.5     & 73.0     & 78.4     & 62.7     & \bf 21.4 & 34.7     & 55.5     & \bf 54.0 & \bf 64.3 & 57.3         \\
		$\delta$-LCS         & \bf 66.8 & \bf 68.3 & 51.6     & \bf 73.3 & 78.4     & 62.9     & 21.3     & \bf 35.4 & \bf 56.2 & 53.5     & 63.9     & \bf 57.4     \\
		$\delta$-LCS$_c$     & 66.6     & 68.1     & \bf 52.0 & 72.5     & \bf 78.6 & 62.0     & 21.3     & 35.0     & 55.3     & 53.9     & 63.5     & 57.2         \\
	\end{tabular}
	\caption{BLEU score (en-fr): \mlevt{3} with the best train/infer pipelines (top) or trained on BM25+LED and inferred on filter-free ED variants.}
	\label{tab:bleu-mlevt-per-domain}
\end{table*}

\begin{table*}
	\centering
	\small
	\begin{tabular}{ll|rrrrrrr|r}
		\hline
		\rowcolor{LightGray}
		model                & k-shot & ECB      & EME      & GNO      & KDE      & News     & PHP      & Ubu      & \textbf{all} \\ \hline\hline
		~~~\textit{test-0.4} &        &          &          &          &          &          &          &          &              \\
		random               & 1      & 26.2     & 27.8     & 37.0     & 27.4     & 24.8     & 19.3     & 42.4     & 29.3         \\
		random               & 3      & 30.7     & 34.7     & 40.1     & 31.8     & 26.9     & 23.4     & 44.2     & 33.1         \\
		LED                  & 1      & 50.7     & 54.4     & 57.6     & 48.0     & 24.2     & 31.3     & 54.5     & 45.8         \\
		LED                  & 3      & 52.7     & 56.7     & 59.0     & 50.2     & 26.1     & 33.0     & 57.2     & 47.8         \\
		LED$_c$              & 3      & \bf 53.6 & \bf 57.6 & \bf 59.9 & \bf 51.3 & 26.3     & 33.0     & 56.7     & \bf 48.3     \\
		BM25                 & 1      & 48.2     & 57.2     & 57.7     & 47.0     & 25.2     & 31.1     & 53.9     & 45.8         \\
		BM25                 & 3      & 52.2     & 58.9     & 59.1     & 49.7     & 27.0     & \bf 33.2 & 56.7     & 48.1         \\
		BM25$_c$             & 3      & 52.7     & 58.7     & 59.4     & 49.5     & \bf 27.3 & 32.4     & 56.6     & 48.1         \\
		$\delta$-LCS         & 3      & 52.8     & 56.6     & 59.0     & 50.9     & 26.4     & 33.1     & 57.8     & 48.1         \\
		$\delta$-LCS$_c$     & 3      & 52.3     & 57.4     & 58.8     & 51.0     & 26.9     & 33.1     & \bf 58.1 & 48.2         \\ \hline\hline
		~~~\textit{test-0.6} &        &          &          &          &          &          &          &          &              \\
		random               & 1      & 26.7     & 26.6     & 33.9     & 25.3     & 21.9     & 18.7     & 43.9     & 28.1         \\
		random               & 3      & 32.1     & 34.1     & 36.5     & 29.3     & 24.3     & 23.2     & 47.5     & 32.4         \\
		LED                  & 1      & 57.9     & 64.7     & 58.7     & 52.9     & 21.5     & 32.8     & 61.4     & 50.0         \\
		LED                  & 3      & 60.7     & 64.6     & 60.3     & 53.8     & 23.4     & 34.8     & 63.2     & 51.5         \\
		LED$_c$              & 3      & \bf 61.5 & 65.1     & \bf 61.0 & 54.1     & 23.7     & \bf 35.2 & 63.5     & \bf 52.0     \\
		BM25                 & 1      & 57.5     & 63.8     & 58.7     & 50.5     & 22.0     & 32.4     & 62.0     & 49.6         \\
		BM25                 & 3      & 59.7     & 65.2     & 60.8     & 53.9     & 24.3     & 35.0     & 64.0     & 51.8         \\
		BM25$_c$             & 3      & 59.6     & 64.9     & 60.3     & 53.7     & \bf 24.1 & 34.1     & 63.0     & 51.4         \\
		$\delta$-LCS         & 3      & 59.6     & \bf 65.3 & 60.8     & 54.5     & 23.3     & 34.8     & 63.0     & 51.6         \\
		$\delta$-LCS$_c$     & 3      & 59.4     & 64.6     & 60.8     & \bf 55.4 & 23.5     & 34.3     & \bf 63.6 & 51.7         \\
	\end{tabular}
	\caption{BLEU score (en-fr): BLOOM inferred on filter-free retrieval settings.}
	\label{tab:bleu-bloom-per-domain}
\end{table*}






\section{Illustration}

\begin{figure*}
	\centering\small
	\begin{tabular}{ll|l}
		source&& Most patients required treatment for their orthostatic hypotension. \\\hline
		NGM+LED && \textit{coverage: } 3 words \\
		&1& Rare: peripheral coldness, \textbf{orthostatic} \textbf{hypotension}\textbf{.} \\
		&2& Entacapone may aggravate levodopa-induced \textbf{orthostatic} \textbf{hypotension}\textbf{.} \\
		&3& - Stalevo may induce \textbf{orthostatic} \textbf{hypotension}\textbf{.} \\\hline
		BM25+LED && \textit{coverage: } 3 words \\
		&1&Hypotension, \textbf{orthostatic} \textbf{hypotension}\textbf{.}\\
		&2&- Stalevo may induce \textbf{orthostatic} \textbf{hypotension}\textbf{.} \\
		&3&Rare: peripheral coldness, \textbf{orthostatic} \textbf{hypotension}\textbf{.} \\\hline
		LED && \textit{coverage: } 3 words \\
		&1&Hypotension, \textbf{orthostatic} \textbf{hypotension}\textbf{.} \\
		&2&- Stalevo may induce \textbf{orthostatic} \textbf{hypotension}\textbf{.} \\
		&3&Rare: peripheral coldness, \textbf{orthostatic} \textbf{hypotension}\textbf{.} \\\hline
		$\delta$-LCS && \textit{coverage: } 6 words \\
		&1&Patients receiving aripiprazole solution \textbf{for} injection should be observed for \textbf{orthostatic} \textbf{hypotension}\textbf{.} \\
		&2&If parenteral benzodiazepine therapy is deemed necessary in addition to aripiprazole solution for injection,\\
		&& \textbf{patients} should be monitored \textbf{for} excessive sedation and for \textbf{orthostatic} \textbf{hypotension} (see section 4.5)\textbf{.} \\
		&3&Hypotension VELCADE \textbf{treatment} is commonly associated with \textbf{orthostatic}/ postural \textbf{hypotension}\textbf{.} \\\hline
		LED$_c$ && \textit{coverage: } 6 words \\
		&1&Hypotension, \textbf{orthostatic} \textbf{hypotension}\textbf{.} \\
		&2&\textbf{Most} \textbf{patients} had relief of symptoms after stopping \textbf{treatment}\textbf{.} \\
		&3&Entacapone may aggravate levodopa-induced \textbf{orthostatic} \textbf{hypotension}\textbf{.} \\\hline
		$\delta$-LCS$_c$ && \textit{coverage: } 5 words\\
		&1&Patients receiving aripiprazole solution \textbf{for} injection should be observed for \textbf{orthostatic} \textbf{hypotension}\textbf{.} \\
		&2&A minority of \textbf{patients} with \textbf{orthostatic} \textbf{hypotension} experienced syncopal events\textbf{.}\\
		&3&If parenteral benzodiazepine therapy is deemed necessary in addition to aripiprazole solution \textbf{for} injection,\\
		&&\textbf{patients} should be monitored \textbf{for} excessive sedation and for \textbf{orthostatic} \textbf{hypotension} (see section 4.5)\textbf{.}
	\end{tabular}
	\caption{\label{tab:illustration-retrievals} Illustration of the variability across some retrieval settings and their respective coverage for a source sentence from EMEA. For each setting, we represent the source-side $3$ best-ranked sentences.}
\end{figure*}

Figure~\ref{tab:illustration-retrievals} illustrates the variability across retrieval settings. We can observe their main characteristics:
\begin{itemize}
	\item Since NGM and BM25 act here as filters, they barely change the retrieved set.
	\item $\delta$-LCS covers more words in the source at the expense of longer sentences.
	\item The first retrieved sentence for the contrastive ranking is always the same as for LED, but the other two are often more diverse, covering more terms.
\end{itemize}

\end{document}